\documentclass[conference]{IEEEtran}
\IEEEoverridecommandlockouts

\usepackage{lineno} % To obtain line numbers
\usepackage{multirow}
\usepackage{ulem}
\usepackage{mathrsfs,amsmath}
\usepackage{dirtytalk}
\usepackage{array}
\usepackage{balance}
\usepackage{flushend}
\newcolumntype{P}[1]{>{\centering\arraybackslash}p{#1}}
\usepackage[detect-none]{siunitx}
\usepackage[belowskip=-5pt,aboveskip=0pt]{caption}

\usepackage{cite}
\usepackage{mathtools}
\usepackage{amsmath,amssymb,amsfonts}
\usepackage{algorithmic}
\usepackage{graphicx}
\usepackage{textcomp}
\usepackage{xcolor}
\def\BibTeX{{\rm B\kern-.05em{\sc i\kern-.025em b}\kern-.08em  T\kern-.1667em\lower.7ex\hbox{E}\kern-.125emX}}

\begin{document}
\title{Causal Inference in Non-linear  Time-series using Deep  Networks and  Knockoff Counterfactuals\\
}

\author{\IEEEauthorblockN{Wasim Ahmad}
\IEEEauthorblockA{\textit{Computer Vision Group} \\
\textit{Friedrich Schiller University Jena}\\
\textit{Jena, Germany}\\
wasim.ahmad@uni-jena.de}
\and
\IEEEauthorblockN{ Maha Shadaydeh}
\IEEEauthorblockA{\textit{Computer Vision Group} \\
\textit{Friedrich Schiller University Jena}\\
\textit{Jena, Germany}\\
maha.shadaydeh@uni-jena.de}
\and
\IEEEauthorblockN{Joachim Denzler}
\IEEEauthorblockA{\textit{Computer Vision Group} \\
\textit{Friedrich Schiller University Jena}\\
\textit{German Aerospace Center (DLR)}\\
 \textit{Institute of Data Science, Jena, Germany}\\
%\textit{Jena, Germany}\\
}
}

\maketitle

\begin{abstract}
Estimating causal relations is vital in understanding the complex interactions in multivariate time series. Non-linear coupling of variables is one of the major challenges in accurate estimation of cause-effect relations. In this paper, we propose to use deep autoregressive networks (DeepAR) in tandem with counterfactual analysis to infer nonlinear causal relations in multivariate time series. We extend the concept of Granger causality using probabilistic forecasting with DeepAR. Since deep networks can neither handle missing input nor out-of-distribution intervention, we propose to use the Knockoffs framework (Barber and Candès, 2015) for generating intervention variables and consequently counterfactual probabilistic forecasting. Knockoff samples are independent of their output given the observed variables and exchangeable with their counterpart variables without changing the underlying distribution of the data. We test our method on synthetic as well as real-world time series datasets. Overall our method outperforms the widely used vector autoregressive Granger causality and PCMCI in detecting nonlinear causal dependency in multivariate time series. 
\end{abstract}
\section{Introduction}

Improving our understanding of cause-effect relationships when analyzing real-world complex systems, e.g.,  economy,  climate- and neuro-sciences, helps in addressing questions like what factors affect our health, the economy, climate, and which actions need to be taken to mitigate the adverse effects and enhance beneficial effects? One of the main challenges in inferring cause-effect relations in such systems is the non-linear dependency among variables, which often leads to inaccurate causal discovery. To mitigate this issue, we propose to use deep recurrent networks in combination with counterfactual analysis to infer nonlinear causal relations in multivariate time series. We apply the concept of Granger causality (GC) \cite{granger1988some} using probabilistic forecasting with deep autoregressive (DeepAR) networks \cite{salinas2020deepar}.   
GC makes use of the following two properties: (i) Temporal precedence: a cause is followed by its effect and (ii) Physical influence: changing cause alters the effects.  It estimates the extent by which a variable $X_i$ helps in the prediction of another variable $X_j$. The impact of the inclusion of $X_i$ in the model is estimated by comparing the prediction of $X_j$ with and without $X_i$.  Due to its practical implementation, GC  has been intensively used in various fields i.e. neuroscience, climatology, finance,  etc.,  for discovering causal graphs in time series \cite{granger1980testing, goebel2003investigating, brovelli2004beta, granger1988some}.

A major shortcoming of Granger causality however is in its assumption of linear dependency in the observational data. Deep networks excel in learning complex and nonlinear relations. In this paper, we use  DeepAR networks \cite{salinas2020deepar} to model non-linear relations in multivariate time series. DeepAR characterizes conditioning on observed variables and extracts hidden features from the time series which helps in dealing with spurious associations that might occur due to confounders such as periodic cycles and trends. However, deep networks can not handle missing input or out-of-distribution data. To deal with this issue, we propose to use the theoretically well-established Knockoffs framework \cite{barber2020robust} to generate intervention variables. The Knockoffs framework was originally developed as a variable selection tool with a controllable false discovery rate in a prediction model. Knockoff samples are statistically null-variables, i.e., independent of the model output, and can be swapped with the original variables without changing the underlying distribution of the data. The in-distribution nature of the knockoffs is important when used to generate counterfactuals for causal discovery because the trained deep networks expect test data to be within the same distribution as training data. Moreover, knockoff samples control false discovery rate in causal inference as it holds as low correlation with the candidate variable as possible.

Causal hierarchy operates in three layers (i) Association (ii) Intervention and (iii) Counterfactual \cite{Pearl2019}. We use counterfactuals where questions like \textit{what if certain thing has been done differently?} Or \textit{is it certain action that causes the change?} are addressed. A counterfactual outcome describes a cause-effect scenario such that if event $X_i$ had not occurred, event $X_j$ would not have happened. Counterfactuals have been used for feature selection, time series explanation, and model interpretability \cite{delaney2021instance, tonekaboni2019explaining}. We generate counterfactual output by substituting one of the observed variables at a time with its knockoff samples and estimate its causal impact by measuring its effect on the forecast error assuming that the used multivariate time series are stationary. Moreover, we assume causal sufficiency, i.e., the set of observed variables includes all possible causes, and there exist no hidden confounders. We compare the causal inference performance of the proposed method with the widely used vector autoregressive Granger causality (VAR-GC) and PCMCI \cite{runge2019detecting} using synthetic as well as real river discharges datasets. We show that our method outperforms these methods in estimating nonlinear causal links. To highlight the advantage of using the Knockoffs framework, we carry out a comparison using different methods for generating counterfactuals. More specifically,  we compare the causal inference results of  when generating counterfactuals using out-of-distribution, a constant value of the distribution-mean, and Knockoffs samples. 
 
The remainder of this paper is organized as follows. In Section \ref{section:relatedwork} of the paper, we cover the related work. Methodological background are presented in Section \ref{section:methods}. In Section \ref{section:causality}, we describe in detail the proposed causal inference method. Experimental results are presented and discussed in Section \ref{section:experiments}. The work is concluded in Section \ref{section:conclusion} of the paper.
%/////////////////////////////////////////////////

\section{Related Work}
\label{section:relatedwork}

Several methods in the literature have addressed the challenge of causal inference in nonlinear multivariate time series. The method of \cite{Tank_2021} uses structured multilayer perceptrons and recurrent neural networks combined with sparsity-inducing penalties on the weights. It monitors the weights assigned to each time series when predicting the target time series.  The authors of \cite{romano2020deep} also proposed a similar solution for causal discovery using deep learning where the architecture of the neural network is partially utilized from data to infer causality among variables. They chose to interpret the weights of a neural network under some constraints. The method trains as many neural networks models as the number of variables in the dataset, where one variable is the target and the rest of the variables are inputs to predict the target. Analyzing the weights of the network reveals whether there is a relationship between a variable and the target: if any of the paths between a variable and the target has a weight value as 0, the target does not have a causal link with that particular variable. However, it is challenging to extract any clear structure of the data from the deep networks that can be used for interpretability \cite{luo2020causal}. Attention-based convolutional neural networks have also been used for causal discovery, where they utilize the internal parameters of the networks to find out time delay in the causal relationship between variables \cite{nauta2019causal}. 

The PCMCI \cite{runge2019detecting} method estimates causality in multivariate time series using either linear or non-linear conditional independent tests. PCMCI works in two steps, first, it determines the causal parents of each variable and then applies conditional independence test to estimate causality between variables. PCMCI has been implemented in \cite{krich2020estimating} to estimate causal networks in biosphere-atmosphere interaction. PCMCI is also used along with dimensionality reduction to infer causality in ecological time series data \cite{krich2021functional} where conditional independence tests are applied on the reduced feature space.

Cause-effect variational autoencoders (CEVAE) \cite{louizos2017causal} estimate non-linear causal links in the presence of hidden confounders by training two separate networks, with and without intervention. This method however estimates the causal link of two variables only. The extended version of CEVAE to time series has been developed by integrating the domain-specific knowledge for inferring nonlinear causality in ecological time series \cite{trifunov2019nonlinear}. Additive models are also used to estimate non-linear interactions in time series, where the past of each series may have an additive non-linear effect that decouples across time series \cite{vantas2020estimating}, however, it may overlook important non-linear relation between variables and therefore may fail to properly identify Granger causal links \cite{Tank_2021}. Counterfactuals are used by \cite{tonekaboni2019explaining} to estimate feature importance. They substitute features to obtain counterfactuals and estimate feature importance by finding their impact on the output. Unlike our approach, it does not estimate causal relations in time series and does not use knockoff in-distribution samples to reduce the false discovery rate. Knockoffs have been used to generate counterfactuals to explain the decision of deep image classifiers \cite{popescu2021counterfactual}. 
To deal with spurious causal links that may result from periodic patterns and trends in time series, the method in \cite{shadaydeh2018causality, shadaydeh2019time}  proposed to use time-frequency causality analysis. However, these methods use the spectral representation of linear VAR-GC and hence may not be able to deal with nonlinear causal relations.

%//////////////////////////////////////////////////
\section{Methodological Background}
\label{section:methods}
In this section we will briefly introduce the necessary background on DeepAR \cite{salinas2020deepar}, the fundamental concept of Knockoffs framework \cite{romano2020deep}, as well as VAR-GC.  
%------------------------------------
\subsection{DeepAR}
DeepAR \cite{salinas2020deepar} is a powerful method mainly used for multivariate non-linear non-stationary time series forecasting. It is based on autoregressive recurrent neural networks, which learn a global model from the history of all the related time-series in the dataset. DeepAR is based on previous work on deep learning for time series data \cite{graves2013generating} and uses a similar LSTM-based recurrent neural network architecture to the probabilistic forecasting problem. Few key attributes of DeepAR compared to classical approaches and other global forecasting methods are: It extracts hidden features in the form of seasonal patterns and trends in the data, and characterizes conditioning on observed variables and extracted features during inference. Moreover, in this probabilistic forecasting method, lesser hand-crafted feature engineering is needed to capture complex, group-dependent behavior.

\begin{figure*}[t]
\begin{center}
  \includegraphics[width=0.98\linewidth]{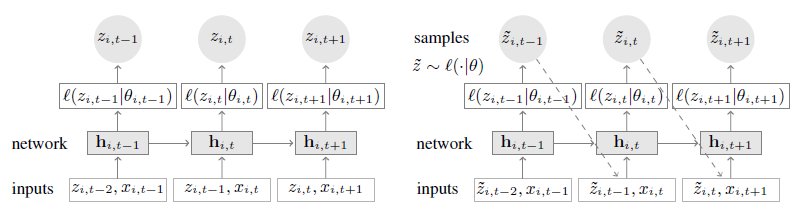}
\end{center}
  \caption{Architecture of DeepAR network showing training (left) and forecast (right) phases \cite{salinas2020deepar}.}

\label{fig:deepar}
\end{figure*}

The training and forecast processes of the DeepAR is shown in Fig. \ref{fig:deepar}. Let $z_i, i = 1, \dots, N$ be the $N-$variate time series. Each time series $z_{i,t}, t = 1, \dots, r$ is a realization of length $r$ real-valued discrete  stochastic process $Z_i, i = 1,\dots, N$. At each time step $t$, the inputs to the network are the observed target value at the previous time step $z_{i,t-1}$ as well as the previous network output $\textbf{h}_{i,t-1}$. The network output $\textbf{h}_{i,t} = h(\textbf{h}_{i,t-1}, z_{i,t-1}, x_{i,t}, \Theta)$ then used to compute the parameters $\theta_{i, t}=\theta(\textbf{h}_{i,t-1}, \Theta)$  of the likelihood $\ell(z|\theta)$, which is used for training the model parameters. For prediction, the past of the time series $z_{i,t}$ is fed in for $t < t_0$, where $t_0$ represents the first value in the forecast horizon, then in the forecast horizon for $t \geq t_0$ a sample $\hat{z}_{i,t} \sim \ell(.|\theta_{i,t})$ is drawn and fed back for the next point until the end of the prediction range $t = t_0 + T$ generating one sample trace. This sampling process is repeated iteratively to generate multiple traces of the variables in the forecast horizon representing the joint predicted distribution.
%%%%%% Figure %%%%%%%%%%%%%%%%%%%%%%

%//////////////////////////////////////////////////////
\subsection{Knockoffs} 
The Knockoffs framework was developed by Barber and Candès in 2015 as a tool for estimating feature importance using conditional independence testing with controllable false discovery rate \cite{Barber_2015}. Given a set of observed variables $Z = (Z_{1}, \dots, Z_{n})$ with known distribution $P_Z$, and a predictive model, the knockoffs of the observed variables, defined as $\widetilde{Z} = (\widetilde{Z}_{1}, \dots, \widetilde{Z}_{n})$, are constructed to be in-distribution null-variables and as decorrelated as possible from the original data. They can be swapped with the original variables without changing the underlying distribution of the data. This is accomplished by ensuring that the correlation between the generated knockoffs is the same as the correlation between the original variables \cite{Barber_2015}. To be null-variables, knockoffs do not contain any information about the target variable. More specifically, knockoff variables should satisfying the pairwise exchangeability condition \cite{Barber_2015}
\begin{equation}
\label{eqn:exchange}
(Z, \widetilde{Z})_{swap(\mathbb{A} )} \overset{d}{=} (\widetilde{Z}, Z)
\end{equation}
for any subset $ \mathbb{A} \subseteq {1,\dots,n}$, here $\overset{d}{=}$ represents equal distributions. 
The $(Z, \widetilde{Z})_{swap(\mathbb{A})}$ is obtained from $(Z, \widetilde{Z})$ by swapping the entries $Z_j$ and $\widetilde{Z_j}$ for each  $j \in \mathbb{A}$. 

The knockoff mechanism can be thought of as generating a probability distribution $P_{\widetilde{Z}|Z} (.|z)$ which is the  conditional distribution of $\widetilde{Z}$ given $Z=z$ chosen such that the obtained joint distribution of $(Z, \widetilde{Z})$ which is equal to
\begin{center}
    $P_{Z}(z) P_{\widetilde{Z}|Z}(\widetilde{z}|z)$,
\end{center}
is pairwise $(z_j, \widetilde{z}_j)$ symmetric and satisfy the exchangeability condition in \eqref{eqn:exchange}. For  Gaussian variables with Gaussian distribution $P_Z = \mathcal{N}_n(\textbf{0}_n, \Sigma)$,  \cite{candes2017panning} shows that knockoffs $\textbf{Z}_{i, *}$ can be drawn from the conditional distribution
\begin{equation}
\label{eqn:condist}
P_{\widetilde{Z}|Z}(.|\textbf{Z}_{i, *}) = \mathcal{N}_n((\textbf{I}_n - S\Sigma^{-1})\textbf{Z}_{i, *}, 2S -S \Sigma^{-1} S)
\end{equation} 
for any fixed diagonal matrix $S$ satisfying $0\leq S \leq 2\Sigma$. This leads to the joint distribution of $(\textbf{Z}_{i, *}, \widetilde{\textbf{Z}}_{i, *})$ being  equal to
\[\mathcal{N}_{2n}
\begin{pmatrix}
\textbf{0}_{2n},
\begin{pmatrix}
\Sigma & \Sigma -S\\
\Sigma - S & \Sigma
\end{pmatrix}
\end{pmatrix},\] which satisfies the pairwise exchangeablity condition in  \eqref{eqn:exchange}.
% These conditions can be formally written as follows.
% \begin{itemize}
% \item[-] Exchangeability: $\forall S\subset 1,\dots,n~~ (Z, \widetilde{Z})_{swap(S)} = (\widetilde{Z}, Z)$, where $ (Z, \widetilde{Z})_{swap(S)}$ is obtained by swapping the features ($Z_j, j \in S$) with their knockoffs $ \widetilde{Z}_j$.
% \item[-] Null variables: $ \widetilde{Z} \bot Y|Z$, which is guaranteed if $\widetilde{Z}$ is generated without looking at the output variable. This condition automatically satisfies in forecasting settings. 
% \end{itemize}

% defined as $Q^{\widetilde{Z}|Z, K}(z|\widetilde{z}, K=k)$. 

% More explicitly: 
% \begin{eqnarray}
%  \large K|Z \sim  P^{K|Z}(k|Z) = \frac{\Large \lambda_k \mathcal{N}(Z;\mu_k , \sum_k)}{\sum_{j=1}^{l} \lambda_j \mathcal{N}(Z;\mu_j, \sum_j)} \nonumber\\
% \widetilde{Z} | Z, K= k \sim Q^{\widetilde{Z}|Z, K}(\widetilde{z}|Z, K=k) = \mathcal{N}(\widetilde{z}; \widetilde{\mu_k}, \widetilde{\sum_k}) 
% \end{eqnarray}
% where 
% \begin{equation}
%     \left\{ \begin{array}{rcl}
% \widetilde{\mu}_k &=& M_k \sum_{k}^{-1} \mu_k + (I_d - M_k \sum_{k}^{-1})Z \\
% \widetilde{\sum}_k &=& 2M_k - M_k \sum_{k}^{-1} M_k
% \end{array}\right. 
% \end{equation}

% as $P^{Z, \widetilde{Z}|K} = Q^{\widetilde{Z}|Z, K} P^{Z|K}$
%//////////////////////////////////////
\subsection{VAR Granger Causality}
GC is based on the idea that cause precedes its effects and can help in their prediction. Here we briefly describe how VAR is used to infer GC. Let $z_i, i = 1,\dots, N$ be the time series of $N$ variables. Each time series $z_{i,t}, t = 1 ,\dots, r$ is a realization of length $r$ real-valued discrete stationary stochastic process $Z_i, i = 1 ,\dots, N$. These $N$ time-series can be represented by a $p_{th}$ order VAR of the form

%%%%%%%% Equation %%%%%%%%%%%%%%
\begin{equation}
\label{eqn:vargc}
  \begin{bmatrix}
z_{1, t} \\ \vdots \\
z_{N, t}
\end{bmatrix}
=
\sum\limits_{m=1}^p A_m
\begin{bmatrix}
z_{1, t-m} \\ \vdots \\
z_{N, t-m}
\end{bmatrix}
+
\begin{bmatrix}
\epsilon_{1}(t) \\ \vdots \\
\epsilon_{N}(t)
\end{bmatrix}
\end{equation}

The residuals $\epsilon_i, i = $ $1, \dots, N$ form a white noise stationary process with covariance matrix $\sum$. The model parameters at time lags $m = 1,\dots, p$ comprise the matrix $A_m = [a_{ij}(m)]_{N\times N}$. Let $\sum_j$ be the covariance matrix of the residual $\epsilon_j$ associated to $z_j$ using the model in \eqref{eqn:vargc}, and let $\sum_{j}^{i-}$ denote the covariance matrix of this residual after missing out the $i_{th}$ row and column in $A_m$. The VAR-GC of $z_i$ on $z_j$ conditioned on all other variables is defined by \cite{geweke1982measurement}.
% %%%%%%%% Equation %%%%%%%%%%%%%%
\begin{equation}
\label{eqn:eqgc}
  \gamma_{i\rightarrow{j}} = \ln\frac{|\sum_{j}^{i-}|}{|\sum_j|}
\end{equation}
%%%%%%%%%%%%%%%%%%%%%%%%%%%%%%%%%%
%//////////////////////////////////////

\section{Causal Effect Estimation using DeepAR and Knockoff Counterfactuals}
\label{section:causality}
\subsection{Granger causality using probabilistic forecasting}
Let $z_i, i = 1,\dots,  N$ be the $N-$variate time series. Each time series $z_{i, t}, t = 1, \dots, r$ is a realization of length $r$ real-valued discrete stochastic process $Z_i, i = 1,\dots, N$. Throughout our study, we make use of the following two assumptions.
\begin{itemize}
    \item[-] Stationarity: $Z_i, i = 1,\dots, N$  is a stationary stochastic process. 
    \item[-] Causal Sufficiency: The set of observed variables $Z_i, i = 1, \dots, N$  includes all of the common causes of pairs in $Z$. 
\end{itemize}
We train and optimize DeepAR by providing multivariate time series $z_{i, t}, i = 1,\dots, N;  t = 1 \dots, r$ as input. For each realization of the time series $z_{i,t}$, we calculate forecast error as mean absolute percentage error (MAPE) using the trained model.
% ///////////Equation///////////////////    
\begin{equation}
\label{eqn:mse}
\mbox{MAPE} = \frac{1}{r}\large\sum_{t=1}^{r} \frac{\mid z_{i, t} - \hat{z}_{i, t}\mid}{\mid z_{i, t}\mid}
\end{equation}
% /////////////////////////////////////
Where $z_{i, t}$ is the actual value of time series $i$ at time step $t$, and $\hat{z}_{i, t}$ is the predicted value  which is defined by the mean of the probability distribution $P(z_{i,t_o:T}|z_{i, 1:t_o - 1})$ at time step \textit{t} in the forecast horizon. To know the cause-effect relationship of $z_i$ on $z_j$,  we  intervene on variable $z_i,i = 1,\dots, N$  with its knockoffs $\widetilde{z}_i$, and notice the influence on the prediction of a target variable $z_j, j\neq i$ by comparing the obtained counterfactual output with the actual output. In analogy to the definition of VAR-GC of $z_i$ on $z_j$ in \eqref{eqn:eqgc}, we define our  metric for estimating  the non-linear Granger causality of  of $z_i$ on $z_j$ conditioned on all other variables by  the causal significance score (CSS$_{i\rightarrow j}$) as follows. 
%-------------------------------------------
\begin{equation}
\label{eqn:css}
    \text{CSS}_{i \rightarrow j} = \ln{\frac{\text{MAPE}_j^i}{\text{MAPE}_j}}  
\end{equation}
%----------------------------------------------
where \text{MAPE$_j$} and \text{MAPE$^i_j$} denote the mean absolute percentage error when predicting  $z_j$ before and after intervention on  $z_i$.
%\\\\\\\\\\\\\\\\\\\\\\\\\\\\\\\\\\\\
\subsection{Counterfactual generation}
In this work, we investigate multiple ways to generate counterfactuals, i.e. distribution mean, out-of-distribution, and knockoff samples as described below. However, we mainly emphasize the use of knockoffs to generate counterfactuals for causal discovery in time series. 
\begin{itemize}
    \item[-] Distribution mean: Here we replace each value $z_{i, t}$ of the time series $z_i$ with the mean  $\overline{z}_i=\frac{1}{r}\sum_{t=1}^{r} z_{i, t}$. 
% ------------------------------------
\item[-]Out-of-distribution: In this type of intervention, the candidate variable $z_i$ that belongs to data distribution $D_i$ is replaced with a variable $\overline{z}$ from another data distribution $\overline{D}$ such that $\overline{D} \overset{d}{\neq} D_i$. Moreover, $\overline{z}_i$ is selected to be  as uncorrelated with the original variable $z_i$ as possible.
% ------------------------------------
\item[-] Knockoffs: The Knockoffs  of the observed variables $Z = (Z_{1}, \dots, Z_{n})$ are constructed to be in-distribution null-variables as decorrelated as possible from the original data. To this end,  we implement  DeepKnockoffs \cite{romano2020deep}, a framework for sampling approximate knockoffs using deep generative models. The knockoff samples are generated using a mixture of Gaussian models as implemented in \cite{gimenez2019knockoffs}. The procedure consists of first sampling the mixture assignment variable from the posterior distribution. The knockoffs are then sampled from the conditional distribution of the knockoffs given the original variables and the sampled mixture assignment such that  the  exchangability condition is  satisfied. We obtain knockoffs   $\widetilde{Z}_{j}, j \subseteq {1, \cdots,n} $, for all variables in the multivariate time series and substitute one variable at a time $Z_j$ with  $\widetilde{Z}_{j}$ for each $j \subseteq {1, \cdots,n}$  for causal inference.
\end{itemize} 
% ------------------------------------
\subsection{Causal hypothesis testing}
We calculate CSS$_{i \rightarrow j}$ using \eqref{eqn:css} by incorporating MAPE$_j$ and MAPE$_j^i$ for each realization $z_{j,t}$ of the  stochastic process $Z_j, j = 1, \dots, N$ before and after intervention on $z_{i,t}$, $i \neq j$. Since we have multiple realizations for each time series, as a result, we get a distribution of values of CSS$_{i \rightarrow j}$ for each pair of time series $z_{i,t}$ and $z_{j,t}$. The  null hypothesis $H_0$:  $z_i$ does not Granger cause  $z_j$, is accepted if the mean of the distribution of  CSS$_{i \rightarrow j}$  is close to zero. In case the mean of the distribution is significantly different from zero, the alternate hypothesis $H_1$: $z_i$ Granger causes  $z_j$ is accepted. The causal relationship for each pair of time series is estimated iteratively and ultimately the full causal graph of the multivariate time series is extracted.

% ///////////////////////////////////////////
\section{Experiments}
\label{section:experiments}
We conducted experiments on synthetic as well as a real datasets. Experiments are performed on multiple realizations of the time series. For synthetic data, we set $r=200$ as the length of each realization $z_{i,t}$ of the time series. The prediction length T is set to 14. We trained our model for 150 epochs. The number of layers and cells in each layer of the network is set to 4 and 40 respectively while dropout is $\numrange{0.05}{0.1}$. We estimate causal link among time series by hypothesis testing on the distribution of CSS$_{i \rightarrow j}$ as explained in Section \ref{section:causality}. To quantify the performance of our method, we calculate  the two metrics: false positive rate (FPR) and F-score where
%-------------------------------------------
\begin{eqnarray}
\label{eqn:fscore}
\text{FPR}&=& \frac{\text{FP}}{\text{FP + TN}} \\
    \text{F-score} &=& \frac{\text{TP}}{\text{TP}+0.5 (\text{FP + FN})}
\end{eqnarray}
%-------------------------------------
Here, TP is the number of causal links that exist in the causal graph and also detected by the method; TN is the number of causal links that neither exist in the causal graph nor detected by the method; FP is the number of causal links that do not exist in the causal graph but detected by the method and FN is the number of the causal links that exist in the causal graph but not detected by the method. 
%\\\\\\\\\\\\\\\\\\\\\\\\\\\\\\\\\\\\\\\\\\\\
\subsection{Synthetic data}
To test the proposed method on synthetic data we generate  time series $X_i(t), i=1, \dots,4$, with seasonal pattern and non-linear relationships among them as follows. This synthetic model simulate a test model of climate-ecosystem interactions as suggested in \cite{krich2020estimating}, where $X_1$ denotes global radiation R$_g$, $X_2$ represents air temperature T$_{air}$ while $X_3$ and $X_4$ denote gross primary production GPP and ecosystem respiration $R_{eco}$ respectively.
% %==========Equations============
\begin{eqnarray}
 X_1(t) &=& \mathcal{N}(0,\, 1) + |cos(2\pi f t)|  \nonumber \\
X_2(t) &=& c_1 X_2(t-\tau_1) + c_2 X_1 (t-\tau_2) + \eta_1(t) \nonumber  \\
X_3(t) &=& c_3 X_1(t-\tau_3) * X_2 (t-\tau_4) + \eta_2(t) \nonumber \\
X_4(t) &=& c_4 X_3(t-\tau_5) * \beta^{\frac{X_2(t-\tau_6)-Q}{10}} + \eta_3(t) 
\end{eqnarray}
% %////////////////////////////////

The coupling coefficients and time lags are represented as $c_1$, $c_2$, $c_3$, $c_4$, $\beta$ and $\tau_1$, $\tau_2$, . . . , $\tau_5$ respectively. The notation $\eta$, termed as \say{intrinsic} or \say{dynamical noise}, which represents data from uncorrelated, normally distributed noise. We set $\eta_1, \eta_2, \eta_3$ as Gaussian with 0 mean and variances [0.30, 0.35, 0.25]. The value for the coupling coefficients $c_1$ to $c_4$ and $\beta$ were set to [0.95, 0.80, 0.50, 0.75] and [0.2, 0.4, ... , 1.0], respectively. The lags were given only integer values in the range [0, 10] and $Q$ is set to 10 which is referred to reference temperature in the test data model used in \cite{krich2020estimating}. The seasonal part of $X_1(t) $ is $|cos(2\pi ft)|$ where $f$ represents the frequency which takes the value 150 and $t$ is the time that varies from 0 to 3000. We keep changing $\beta$, the coefficient of non-linear relationship between $X_2$ and $X_4$, in the test model every simulation while rest of the parameters are kept intact. We are interested to see the consistency and causal estimation capability of our method in comparison with PCMCI and VAR-GC when $\beta$ is increased.

%////////////////////////////////////////
\subsection{Real data}
%Ecological Dataset
% We used real ecological time series “FLUXNET2015”, acquired using eddy covariance technique. The data set provides data on $CO_2$, water, and energy exchange between the biosphere and the atmosphere, from 212 sites around the globe. This data set is used in many applications i.e. remote sensing studies, and development of ecosystem and Earth system models etc. It contains variety of measured and derived time-series, however, based on research literature, time series like $T_{air}$, \textit{GPP}, precipitation \textit{PPT}, vapor pressure deficit \textit{VPD} and \textit{$R_{eco}$} are predominantly studied and modelled in order to know their interaction with each other.
% River Discharge Data 
As a real-world application of our method, we carry out causal analysis of average daily discharges of rivers in the upper Danube basin, provided by the Bavarian Environmental Agency \footnote{https://www.gkd.bayern.de}. We use measurements of three years (2017-2019) from the Iller at Kempten $K_t$, the Danube at Dillingen $D_t$, and the Isar at Lenggries $L_t$ as considered by \cite{gerhardus2020high}. While the Iller flows into the Danube upstream of Dillingen with the water from Kempten reaching Dillingen within a day approximately, the Isar reaches the Danube downstream of Dillingen. Based on the stated scenario, we expect a contemporaneous link $K_t \rightarrow{D_t}$ and no direct causal links between the pairs $K_t$, $L_t$ and $D_t$, $L_t$. Since all variables may be confounded by rainfall or other weather conditions, this choice allows testing the ability of the methods to detect and distinguish directed and bidirectional links.
%//////////////////////////////////////////////

% %-- -----Reco modelling results -------------
% \begin{table}[h]
% \small
% \centering
% \caption{Causal effect of ecological system variables on $R_{eco}$ time series}
% \label{tab:results}
% \begin{tabular}{l l l}
% \hline
% \thead{\textbf{Target}} &  \thead{\textbf{Causality}} & \thead{\textbf{CSS}} \\
% \hline
% $R_g \rightarrow{} R_{eco}$ & Yes & - \\
% \hline

% $T_{air} \rightarrow{} R_{eco}$ & Yes & -  \\
% \hline

% GPP $\rightarrow{} R_{eco}$ & Yes & - \\
% \hline

% VPD $\rightarrow{} R_{eco}$ & Yes & - \\
% \hline

% PPT $\rightarrow{} R_{eco}$ & No & - \\
% \hline
% \end{tabular}
% \end{table}
%//////////////////////////////////

\subsection{Results}
\begin{figure}[t]
\begin{center}
   \includegraphics[width=0.95\linewidth]{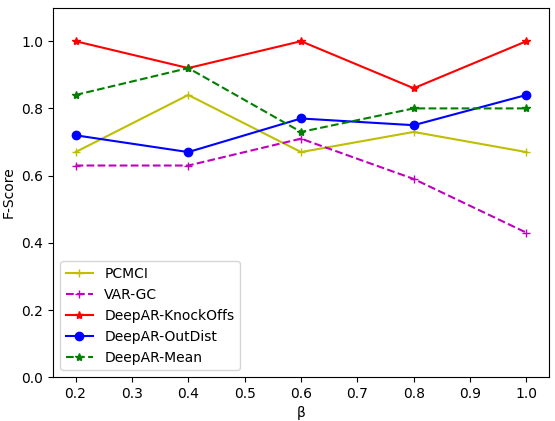}
\end{center}
\caption{F-score of given methods in response to increase in the coefficient of non-linear relationship $\beta$ between $X_2$ and $X_4$.}
\label{fig:fscorecomp}
\end{figure}

\begin{figure}[t]
\begin{center}
   \includegraphics[width=0.95\linewidth]{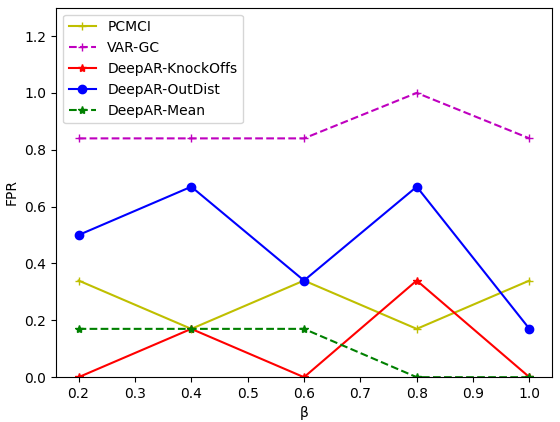}
\end{center}
\caption{False positive rate (FPR) of all methods for increasing $\beta$ between $X_2$ and $X_4$ in the synthetic data model.}
\label{fig:fpr}
\end{figure}

% Causal estimation of the variables in test model dataset are shown in Table \ref{tab:results}. The dataset has four time series with a known causal dependency as given in the test model. There is confounding scenario in case of \textit{T} as a common cause for \textit{GPP} an $R_{eco}$. As mentioned earlier, we use three types of counterfactuals to estimate GC using trained DeepAR. We show results for our method applying three types of counterfactuals and compare it with VAR-GC and PCMCI. Our method outperform these two widely used methods because of the ability to deal with confounder and seasonality. $R_g_{obs}$ has a daily seasonality. Our method handles seasonality because it extract seasonal information from the time series as derived feature and incorporate conditioning on these derived features and covariates in the forecasting settings. 
For synthetic data model, we demonstrate results in Fig. \ref{fig:fscorecomp} which shows the F-score values for PCMCI, VAR-GC, and DeepAR with all three types of substitution methods represented as DeepAR-Knockoffs, DeepAR-OutDist, DeepAR-Mean. The figure illustrates the performance of these methods in response to an increase in the coefficient of non-linearity $\beta$ between $X_2$ and $X_4$. VAR-GC assumes linearity and hence yields a lower F-score as the dataset contains non-linear relations among variables in the synthetic data model. PCMCI however has better results compare to VAR-GC because of using a non-linear method, Gaussian process distance correlation (GDPC), for testing conditional independence. DeepAR-Knockoffs, yields better results in terms of F-score compare to VAR-GC and PCMCI. DeepAR-OutDist suffers from a high false discovery rate because of using out-of-distribution variables as substitution and hence yield lower F-score throughout the experiment. DeepAR-Mean performs better than DeepAR-OutDist but falls behind DeepAR-Knockoffs. The plot in Fig \ref{fig:fpr} shows FPR in response to increase in the non-linearity coefficient $\beta$. It can be seen that VAR-GC yields a high FPR due to its assumption of linearity. Besides that, DeepAR-OutDist also produces a high number of false positives, because outside-distribution samples introduce bias in the model and hence results in detection of false causal links. PCMCI, DeepAR-Mean, and DeepAR-Knockoffs however have better false discovery rates.

%/////////////////////////////////////
%////////////////////////////////////////
%\\\\\\\\\\\\\\\\\\\\\\\\\\\\\\\
For real data, the contemporaneous link $K_t \rightarrow{D_t}$ is correctly detected by DeepAR-Knockoffs, VAR-GC and PCMCI as shown in Table \ref{tab:riverdischarge}. Our method does not report any causal link $D_t \rightarrow{K_t}$ however VAR-GC and PCMCI wrongly finds $D_t \rightarrow{K_t}$. PCMCI and VAR-GC detected a causal link $K_t \rightarrow{L_t}$ and $L_t \rightarrow{K_t}$ which may be because of the weather acting as a confounder. DeepAR-Knockoffs does not report a causal link between $K_t$ and $L_t$ in any direction as the method has the capability of handling confounders, such as seasonal patterns,  in a better way. Our method wrongly finds a causal link $L_t \rightarrow{D_t}$, however, VAR-GC and PCMCI do not detect such a causal relationship. Due to extreme events in the river discharges data caused by heavy rainfall, the assumption of stationarity is expected to be violated, which is why VAR-GC and PCMCI report these false links. In comparison DeepAR-Knockoffs correctly discovers the expected links in the river discharges data, except a single false link.

%////////////////////////////////////////

\begin{table}[ht]
%\small
\centering
\caption{Expected and detected causal links in river discharges data by VAR-GC, PCMCI and DeepAR-Knockoffs}
\begin{center}
\begin{tabular}{|P{1.3cm}|P{1.3cm}|P{1.3cm}|P{1.3cm}|P{1.3cm}|}
    \hline
    Causal links & Expected & VAR-GC & PCMCI & DeepAR-Knockoffs\\
    \hline
    $K_t \rightarrow D_t$ & Yes & Yes & Yes & Yes \\
    \hline
    
    $K_t \rightarrow L_t$ & No & Yes & Yes & No \\
    \hline
    
    $D_t \rightarrow K_t$ & No & Yes & Yes & No \\
    \hline
    
    $D_t \rightarrow L_t$
    & No & Yes & No & No \\
    \hline
    
    $L_t \rightarrow K_t$
    & No & Yes & Yes & No \\
    \hline
    
    $L_t \rightarrow D_t$
    & No & No & No & Yes \\
    \hline
\end{tabular}
\end{center}
\label{tab:riverdischarge}
\end{table}
%///////////////////////////////////////////////
%-----------------------------------------------

\section{Conclusion}
\label{section:conclusion}
We proposed a novel method for inferring cause-effect relations in non-linear multivariate time series. Since deep networks can not handle out-of-distribution intervention, we proposed to use probabilistic forecasting with DeepAR in combination of knockoffs-based counterfactuals for estimating nonlinear Granger causality. We applied our method on synthetic and real river discharges datasets. Our results confirm that using knockoff samples to generate counterfactuals yields better results in comparison  to using the mean- and out-of-distribution as intervention methods. Results also indicate that the proposed method outperforms VAR-GC and PCMCI methods in estimating non-linear relations and dealing  with confounders. It should be noted however that the better performance of the proposed method comes along with the higher computational load associated with using deep networks specially for large multivariate time series.  

\section{Acknowledgment}
This work is funded by the Carl Zeiss Foundation within the scope of the program line \say{Breakthroughs: Exploring Intelligent Systems} for \say{Digitization — explore the basics, use applications} and the DFG grant SH 1682/1-1.  

%/////////////////////////////////

\bibliography{./IEEEICMLA}

\end{document}